# An OvS-MultiObjective Algorithm Approach for Lane Reversal Problem

E. G. Baquela
GISOI - FRSN
Universidad Tecnológica Nacional
San Nicolás de los Arroyos, Argentine
ebaquela@frsn.utn.edu.ar

A.C. Olivera
DCEyN,UACO
Universidad Nacional de la Patagonia Austral
Santa Cruz, Argentine
aolivera@uaco.unpa.edu.ar

*Abstract*- **The lane reversal has proven to be a useful method to mitigate traffic congestion during rush hour or in case of specific events that affect high traffic volumes. In this work we propose a methodology that is placed within optimization via Simulation, by means of which a multi-objective genetic algorithm and simulations of traffic are used to determine the configuration of ideal lane reversal.**

*Keywords—optimization; simulation; traffic; genetic algorithms; lane reversal*

## INTRODUCTION

The Lane Reversal has proven to be a very effective method to solve traffic congestion problems when there appear what bibliography calls traffic waves (temporary changes in the dynamics of vehicle traffic that unbalance the flow of traffic in a network) [1]. These traffic waves are generated, in most cases, by the dynamics of traffic during rush hour, by the flow generated after an event (a show, for example) or by emergencies [2].

There exists a significant amount of works in which this topic is dealt with. Reference [3] analyzes possible applications of lane reversal and [4] formalizes the problem of selection of lanes to be reversed with a nonlinear programming model. Reference [2] analyzes the status of 53 implementations distributed worldwide. Reference [5] studies the characteristics of traffic systems submitted to lane reversal; [6] analyzes the impact of lane reversal according to the characteristics of traffic routes and of the vehicle flow pattern. On the other hand, [1] analyzes the bibliography that is available and, detecting that investigations deal with lane reversal in isolation regarding the rest of traffic policies, proposes a multi-objective mixed linear programming model that integrates lane reversal and traffic lights coordination, in order to optimize the system comprehensively.

Despite the vast amount of investigations about this topic, the selection of lanes to be reversed is carried out by means of a mathematical programming model, locally to the routes under study, without taking into account either the effect over the whole traffic system or the dynamic behavior of such systems (for example, spontaneous blockages due to relative differences of speed). Reference [7] uses formalism based on the Optimization via Simulation (OvS) to deal with the related problem of traffic contraflow, using Genetic Algorithms in the microscopic simulation and optimization cycle for the evaluation of solutions, but the local environment of the problem under study is still being restricted.

In this work we propose an optimization model based on OvS that uses a Multi-Objective Genetic Algorithm and microscopic simulations of traffic to determine which lanes are going to be reversed, evaluating the effect of each alternative over the whole system and taking into account their dynamic effects.

## DEFINITION OF THE PROBLEM

Given a traffic network T, with a distribution of the traffic flow F (traffic flow in normal conditions) associated to a number of I vehicles, the direction of lanes and roads that belong to set C is intended to be modified in such a way as to absorb the variation in the traffic flow distribution linked to a specific pattern of circulation P, mitigating the impact in the performance of the traffic system.

Our decision variables (called X) are associated with the decision of reversing or not the traffic flow in a particular lane. Therefore, we have N binary variables, being N the number of lanes that are available for reversal. For coding purposes, we consider that value "1" represents directionality reversal and "0" represents no reversal.

The definition of objective function admits many points of view according to the nature of the disturbance. On one hand,

we want the traffic wave impact to be as small as possible over the vehicle flow. For this reason, we choose to maximize the average speed of the vehicles that circulate by the system (1). On the other hand, the lane reversal involves some side effects. Drivers may initially be out of the habit of driving by a lane that has changed its directionality, with the subsequent difficulty to rearrange their road map. Therefore, it is desirable to have as few changes in the network as possible. So, we will mitigate the number of changes to the system as well (2). Then we have a multi-objective nonlinear programming problem with binary decision variables, where each one represents the decision of reversing traffic directionality in a particular lane.

$$\text{Max } Z1 = \Sigma(\text{Speed}_i)/\ I \qquad (1)$$

$$\text{Min } Z2 = \Sigma X \qquad (2)$$

### RESTRICTIONS TO THE SYSTEM

The main restriction to the selection of lanes to be reversed consists in defining in which lanes it is possible to change directionality. This is defined by choosing which lanes are going to be represented by decision variables and which ones are not. Restrictions related to lanes capacity, interconnection of them and other network characteristics are defined implicitly in the simulation model, being left for explicit definition only the restrictions related to lane reversal policies. For example, lanes in which directionality reversal are mutually exclusive, lanes with direction dependency, etc. Due to the adopted formalism, these restrictions are modeled as binary restrictions and they do not represent great obstacles.

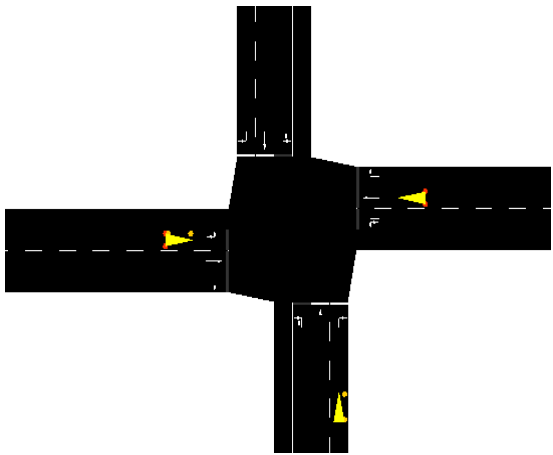

Fig. 01: Traffic Simulation (cross of two avenues)

### MODELING BY MULTI-OBJECTIVE GENETIC ALGORITHMS

For this problem, the chosen modeling and resolution algorithm was a multi-objective genetic algorithm (MOGA). MOGAs have proven to be very useful to optimize multi-objective nonlinear systems, generating solutions that adjust to Pareto boundary of the two objectives. Besides, the use of binary genetic algorithms is presented as a natural way to model the decision variables of this problem, owing to the fact that each bit of a particular chromosome represents the reversal or not of the lane related to it. Then, given a predefined traffic network, there exists a correlation 1:1 between each chromosome and each variation of the original structure of the network.

The MOGA scheme used is an adaptation of [8]. In our research, elitism was not used; crossing in a point was generated by binary competence and a simple mutation scheme. As regards crossing, crossing schemes of 10, 25 and 50% of the chromosomes were tested.

The mutation that was used was of 0,5, 0,1 and 0,15 in the chromosome level. The bit to be mutated was selected with a uniform probability.

### SYSTEM SIMULATION

The evaluation of the target function Z1 is carried out by means of traffic microscopic simulations. In a microscopic simulation, the individual behavior of each vehicle is simulated, arising the global dynamics of the system from the interactions among them. These interactions are in general simple, since vehicles are modeled by means of a pattern of the type "Car-Following", in which the speed of a vehicle depends on its own speed and on the immediately preceding car's speed. In Fig. 01, a traffic simulation is shown.

With the aim of evaluating the target function Z1, it is necessary to configure the simulation model according to the characteristics of the solution to be evaluated. Firstly, it is necessary to re-structure traffic network connections so that they can be representative of the new distribution of directions. Then, starting from a fixed traffic flow definition (i.e. a number of vehicles that start in an origin i towards a destination j), it is necessary to translate such flows into an individual route for each vehicle. Afterwards, the simulation is carried out during a representative time horizon to the traffic wave that was studied, giving as a result the total time of the travel for each vehicle. Finally, the average travel time is calculated as the arithmetical average of the previous values.

The simulation admits the definition of various additional parameters, being the degree of random variation of vehicles' speed the most important one. This parameter has to be calculated on the basis of samples of the system under analysis. However, it has the effect of allowing us to simulate traffic congestion in a realistic way, not because of saturation of the network's capacity but due to speed variations among vehicles.

When simulated all traffic network, the target function is not limited to evaluate the environment of vehicles directly affected by the traffic wave but the impact over the total number of vehicles that circulate by the system.

### COMPLETE OPTIMIZATION PROCESS

So, the complete process of OvS can be summarized into the following steps [8]:

1) Generation of the initial population of solutions to be evaluated.

2) Evaluation of each solution.

2.1) Generation of a traffic network for each scenario (or solution).

2.2) Calculation of individual routes of each vehicle for each scenario.

2.3) Simulation of each scenario.

3) Order of solutions for the criterion of non-dominant solutions.

4) Selection of two parents for crossover using the binary tournament selection method.

5) Crossover

6) Generation of mutation with probability M for each offspring.

7) Incorporation of the offspring to the new population, repetition of 4 to 6 until generating a new population with the same size as the initial one.

8) Repetition of steps 2 to 7.

In an attempt not to waste computer time, we keep a record of the solutions that were evaluated in order to avoid repeating unnecessary simulations.

### EXPERIMENTS AND RESULTS

In order to test the methodology, rectangular traffic networks were used, with pre-generated traffic flows and traffic waves randomly generated. In Fig. 02 a part of traffic net is shown.

Fig. 03 shows the Pareto frontier for one example, sampling the best solution in different generations. It can be seen that an increase in the amount of change generates an interesting restructuring of the network traffic, which increases average speeds in the system.

In Fig. 04 the evolution of the average speed is displayed when the number of changes is restricted to a fixed value. A large number of generations needed for the algorithm to detect a set of lanes directly related to the traffic wave. Selecting initial solutions invest close lanes to traffic flow generated by the waves; it is possible to accelerate the process.

### CONCLUSIONS

The developed lane reversal optimization model has shown satisfactory results on the tests that were carried out, and it proved to be a useful alternative close to reality to evaluate lane reversal policies.

Moreover, with reference to the lane reversal problem itself, the apparently opposite objectives of mitigating the impact over average speed and generating as few changes as possible are not necessarily opposite objectives.

## *Acknowledgment*

The work of E.G. Baquela was supported by the Universidad Tecnológica Nacional, under PID 816.

The work of A. C. Olivera was supported by the Consejo Nacional de Investigaciones Científicas y Técnicas of Argentina and the ANPCyT under Contract PICT-2011-0639.

## References

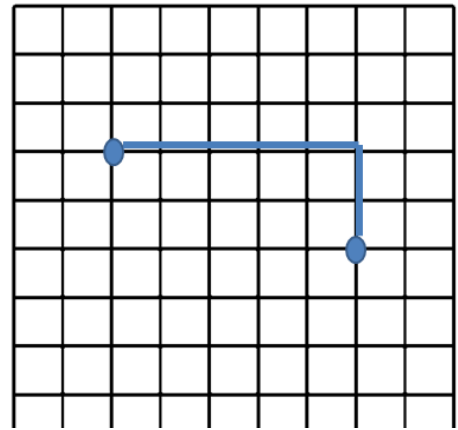

Fig. 02: Traffic Net (two nodes and one of possible routes between them)

J. Zhao, W. Ma, Y. Liu, and X. Yang, "Integrated design and operation of urban arterials with reversible lanes," Transportmetrica B: Transport Dynamics, vol. 0, pp. 1-21, May 2014.

R. Kishore Kamalanathsharma, "Reversible Lanes: State of Implementation on a Global Level.", Annual ATES Student Paper Competition, March 2011.

Z. Bede, G. Szabó, and T. Péter, "Optimalization of road traffic with the applied of reversible direction lanes," Transportation Engineering , vol. 38-1, pp. 3-8, 2010.

Z. Bede, and T. Péter, "The mathematical modeling of reversible lane system," Transportation Engineering , vol. 39-1, pp. 7-10, 2011.

L. Lambert, and B. Wolshon, "Characterization and comparison of traffic flow on reversible roadways," Journal of Advanced Transportation , vol. 44-2, pp. 113-122, April 2010.

W. Cheng, X. Li, and S. Liu, "Research on the traffic control and organization coordinate applications of reversible lanes," 3rd International Conference on Awareness Science and Technology (iCAST) , pp. 72-75, September 2011.

A. Karoonsoontawong, "Genetic Algorithms for dynamic contraflow problem," Journal of Society for Transportation and Traffic Studies, vol.1-2, pp. 1-17, June 2010.

H. W. Ding, L. Benyoucef, and X.Xie, "Stochastic multi-objective production-distribution network design using simulation-based optimization," International Journal of Production Research, vol.47-2, pp. 479-505, 2009.

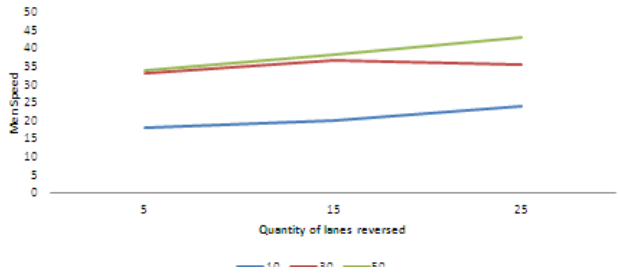

Fig. 03: Pareto Frontier from generations 15, 30 and 50.